\documentclass[letterpaper, 10 pt, journal, twoside]{IEEEtran} 
\usepackage{enumerate}
\usepackage{bm}
\usepackage{multirow}
\usepackage{graphicx}
\usepackage{float}
\usepackage{subfig}
\usepackage{wrapfig}
\usepackage{array}
\usepackage{color}
\usepackage{amsmath}
\usepackage{cite}
\usepackage{fancyhdr} 
\IEEEoverridecommandlockouts                              

\begin{document}
\title{
Planning Irregular Object Packing via Hierarchical Reinforcement Learning
}

\author{Sichao Huang, Ziwei Wang, Jie Zhou, and Jiwen Lu

}

\markboth{IEEE Robotics and Automation Letters. Preprint Version. Accepted October, 2022}
{Huang \MakeLowercase{\textit{et al.}}: Planning Irregular Object Packing via Hierarchical Reinforcement Learning} 

\maketitle
\begin{abstract}
Object packing by autonomous robots is an important challenge in warehouses and logistics industry. Most conventional data-driven packing planning approaches focus on regular cuboid packing, which are usually heuristic and limit the practical use in realistic applications with everyday objects. In this paper, we propose a deep hierarchical reinforcement learning approach to simultaneously plan packing sequence and placement for irregular object packing. Specifically, the top manager network infers packing sequence from {six principal view heightmaps} of all objects, and then the bottom worker network receives heightmaps of the next object to predict the placement position and orientation. The two networks are trained hierarchically in a self-supervised Q-Learning framework, where the rewards are provided by the packing results based on the top height , object volume and placement stability in the box. The framework repeats sequence and placement planning iteratively until all objects have been packed into the box or no space is remained for unpacked items. We compare our approach with existing robotic packing methods for irregular objects in a physics simulator. Experiments show that our approach can pack more objects with less time cost than the state-of-the-art packing methods of irregular objects. {We also implement our packing plan with a robotic manipulator to show the generalization ability in the real world.}
\end{abstract}

\begin{IEEEkeywords}
Manipulation planning, reinforcement learning, robotic packing.
\end{IEEEkeywords}

\footnote{Manuscript received: July 3, 2022; Revised: September 29, 2022; Accepted: October 27, 2022.}
\footnote{
	This paper was recommended for publication by Editor Hong Liu upon evaluation of the Associate Editor and Reviewers' comments. This work was supported in part by the National Natural Science Foundation of China under Grant 62125603 and Grant U1813218, and in part by a grant from the Beijing Academy of Artificial Intelligence (BAAI). \textit{(Corresponding author: Jiwen Lu.)}}\\
\footnote{Sichao Huang, Ziwei Wang, Jie Zhou, and Jiwen Lu are with the Beijing National Research Center for Information Science and Technology (BNRist), the Department of Automation,
	Tsinghua University, Beijing 100084, China (e-mail: huangsc20@mails.tsinghua.edu.cn; wang-zw18@mails.tsinghua.edu.cn; jzhou@tsinghua.edu.cn; lujiwen@tsinghua .edu.cn).}\\
\footnote{Digital Object Identifier (DOI): see top of this page.}

\IEEEpeerreviewmaketitle

\section{INTRODUCTION}

Due to the rapid development of the E-commerce and labor shortages, robotic packing has attracted more and more interest in warehouse automation in recent years~\cite{Wang2019,Tanaka2020}. Compared to conventional methods, automatically packing with robotic arms brings significant benefits including higher efficiency, increased space utilization and lower accident rates. An automatic packing system consists of environment perception, packing planning and robotic manipulation modules. Among these above modules, packing planning module generates the optimal packing sequence and placement for the arm manipulator according to the object information received from environment sensors. Hence, fully utilizing the packing space with the optimal packing plan is desirable for effective robotic packing systems.
   
Packing planning is an NP-hard combinatorial optimization problem with high complexity. In order to efficiently generate the optimal sequence and placement of objects, heuristic methods \cite{Wang2019,Karabulut2004,Zhao2021Advanced} with the greedy objective minimize the object stack heights in the packing boxes. Since the greedy search results in sub-optimal solution and high computational cost for object arrangement, data-driven methods \cite{Hu2017,Tanaka2020,Zhao2021} employ the reinforcement learning framework for bin packing. However, the objects for packing in realistic applications are usually irregular. Conventional learning based methods for cuboid packing fail to perceive fine-grained information from the high-resolution visual clues of irregular objects, and their practicality to pack objects with various shapes and appearance is still limited.
   
\begin{figure}[t]
    \centering
    \includegraphics[width=\linewidth]{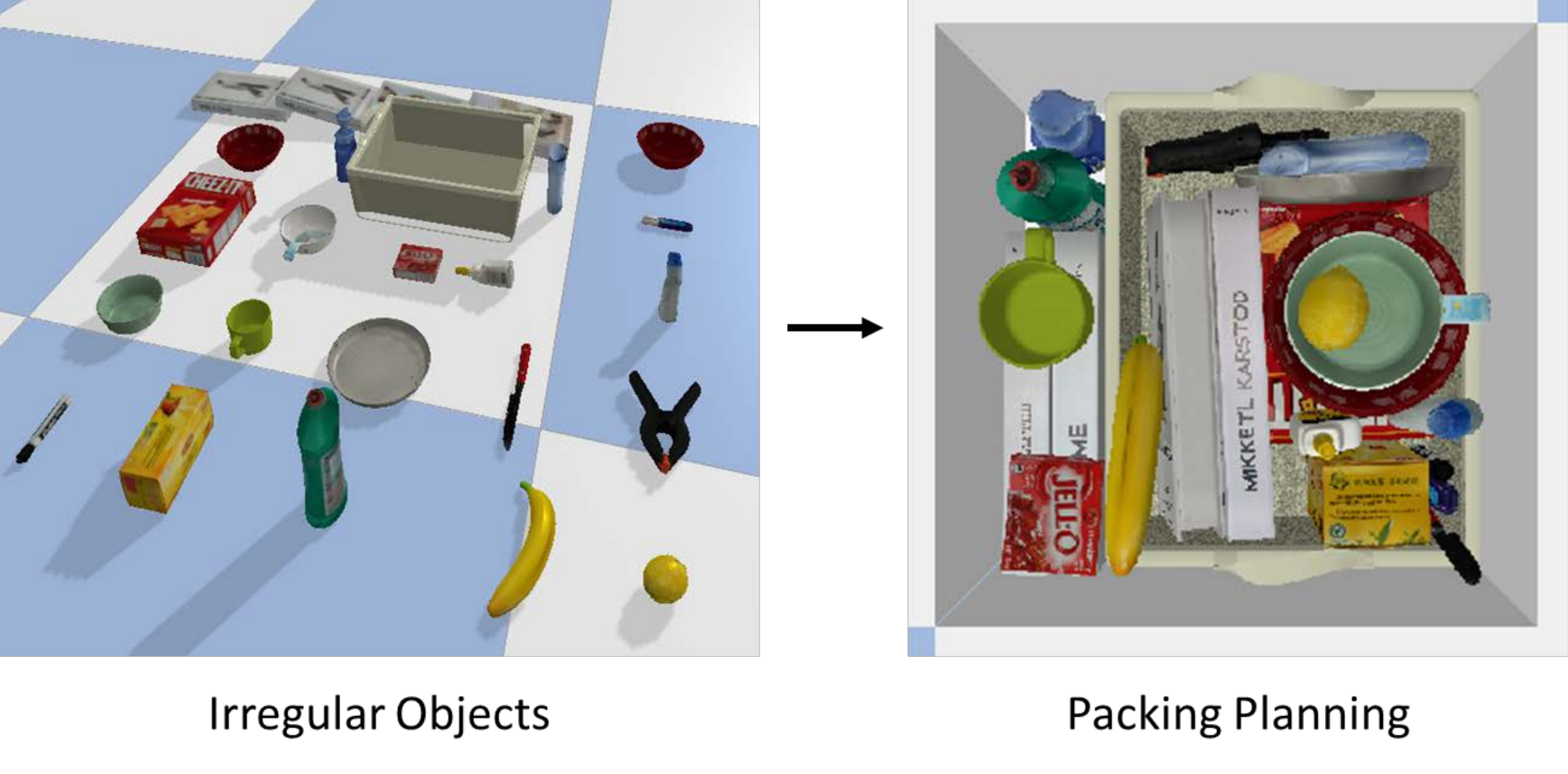}
    \caption{An example of packing planning for irregular objects. Objects should be packed into the empty box with optimal sequence and placement.}
    \vspace{-0.5cm}
    \label{fig:overview}
    \end{figure}
   In this paper, we propose a deep hierarchical reinforcement learning method to simultaneously plan sequence and placement for irregular object packing. {With 3D object models,} our method extracts fine-grained information from the high resolution visual clues to enhance the practicality for irregular object packing, and effectively search for the optimal solution for object arrangement in a data-driven manner. More specifically, the high-level manager in hierarchical reinforcement learning chooses the next object from all remaining instances according to the {heightmaps of six principal views (front, rear, left side, right side, top and bottom)} via convolutional networks. The low-level worker yields score prediction for different placement locations and orientations of the selected object, and picks the placement with highest score based on the heightmaps of the selected object and the packing box. The packing box is initialized to be empty for irregular object packing, and object instances are packed into the box through iterative sequence and placement planning. The packing process stops until all objects are contained in the box or no space is remained for unpacked objects. With the hierarchical architectures, the extremely large search space of object sequence and placement for packing arrangement is efficiently explored with optimal solution generation. The manager and worker networks are jointly trained in a self-supervised Q-Learning framework, where the rewards are acquired from box space utilization and the placement stability. Fig. \ref{fig:overview} demonstrates an example of packing planning for irregular objects. We evaluate our method in the simulated environment by packing random item sets from real-world object scanning. Experiments show that our method outperforms the state-of-the-art packing planning methods with respect to the number of packed objects in given container and the latency for plan generation. Moreover, the increase in effectiveness and efficiency for irregular object packing is more significant for objects with smaller size and thickness. {We also deploy our packing planning method in a robotic manipulator to show the generalization ability in the real world.}

\begin{figure*}[t!]
\centerline{\includegraphics[width=1.03\linewidth]{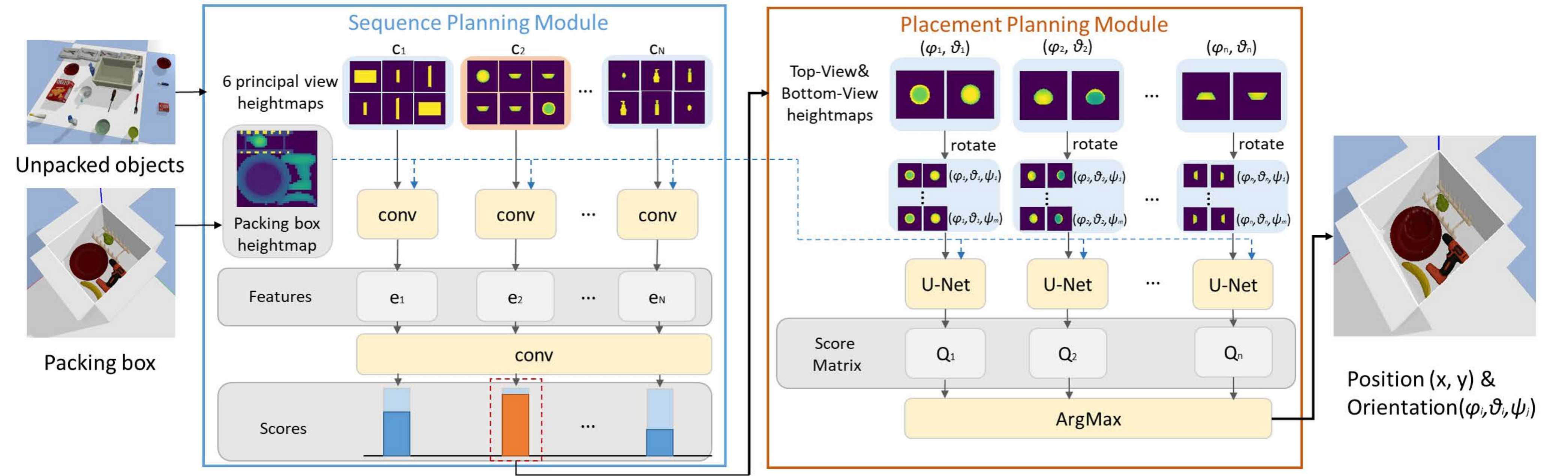}}
\caption{The overall pipeline of the proposed method which consists of the sequence planning module and the placement planning module. The sequence planning module predicts the next object to pack based on the {six principal view heightmaps} of all unpacked objects and the box heightmap, and the placement planning module generates the optimal placement position and orientation for the selected object by its top-view and bottom-view heightmap and the current object arrangement in the packing box.}
\vspace{-0.5cm}
\label{fig:pipeline}
\end{figure*}

\section{Related Work}

In this section, we briefly review two related topics: 1) Packing planning and 2) hierarchical reinforcement learning.

\label{sec:citations}
\subsection{Packing Planning} 
Packing planning is a classic NP-hard combinatorial optimization problem \cite{Chazelle1989}. Most research on packing planning focuses on packing 2D or 3D rectangle objects, and existing methods can be divided into two categories: heuristic~\cite{Iori2021,che2021machine,abdel2018improved} and data-driven approaches~\cite{Ali2022,zhang2021attend2pack,mazyavkina2021reinforcement}. Heuristic methods generate the sequence and placement of objects by greedy objective regarding height, surface area and volume. Jangiti \emph{et al.} \cite{Jangiti2018} proposed a deterministic heuristic based on angle-occupying placement with maximum fit degree. El \emph{et al.} \cite{El2019} modified squirrel search algorithm for improved best fit heuristic to solve large-scale instances within a reasonable time. To address the problem of sub-optimal solution and high computational cost in heuristic methods, data-driven methods \cite{Ali2022} explored the large search space by efficiently interacting with the environment. {Zhang \emph{et al.}\cite{zhang2021attend2pack} encoded the input with self-attention and decoded the sequence and placement with reinforcement learning for bin packing problem.} Hu \emph{et al.} \cite{Hu2017} generated the optimal bin packing sequence via deep reinforcement learning, and Tanaka \emph{et al.} \cite{Tanaka2020} simultaneously planned picking and placing for efficient transportation of cuboids.
Nevertheless, irregular objects instead of cuboids are usually packed in realistic applications, and yielding optimal object sequence and placement for irregular objects according to the visual appearance has aroused extensive interest. Meta-heuristic methods including Tabu Search (TS)~\cite{Bortfeldt1998} and Guided local search (GLS)~\cite{Faroe2003} begin with randomly placed objects and iteratively minimize the objective function by moving objects with collision prevention~\cite{Zhao2021Advanced}. Constructive positioning heuristic methods containing Empty Maximal Space (EMS)~\cite{Ramos2016}, Maximum Contact Area (MCA)~\cite{Zhao2019} and Heightmap Minimization (HM)~\cite{Wang2019} pack objects into empty container in the order determined by heuristic rules. {Goyal \emph{et al.}\cite{goyal2020packit} proposed PackIt for irregular object packing that decides sequence with deep networks and selects position and orientation with heuristics.} However, the greedy search for irregular object packing results in sub-optimal solution and high computational cost in object arrangement, which significantly limits the practicality in realistic applications. 

\subsection{Hierarchical Reinforcement Learning}

Hierarchical reinforcement learning aims to effectively explore the large search space by decomposing the complex selections into various hierarchies, because the space shrinks significantly for search efficiency enhancement \cite{kulkarni2016hierarchical,vezhnevets2017feudal,nachum2018data}. Early attempts \cite{sutton1998intra} presented the option framework where the agent in the top level assigned the goal for the search process and the agent in the bottom level achieved the goal by selecting the optimal action primitives. Recently, hierarchical reinforcement learning has been extended to a wide variety of tasks including machine learning \cite{florensa2017stochastic,eppe2022intelligent,wang2020learning}, computer vision \cite{wang2018video,zhao2017multiresolution} and robotic manipulation \cite{kim2021landmark,yang2021hierarchical,lampinen2021towards}. For the first regard, Wang \emph{et al.} \cite{wang2020learning} employed the top-level agent to partition channels of binary networks and utilized the bottom-level agent to discover the optimal fine-grained channel-wise interaction in each partition, so that the network capacity is strengthened with reduced the quantization errors. Wan \emph{et al.} \cite{wan2021reasoning} proposed a hierarchical reinforcement learning framework for encoding historical information and learning structured action space to learn chains of reasoning from a knowledge graph. In computer vision, Wang \emph{et al.} \cite{wang2018video} modeled video caption generation as a two-level search problem, where the manager selected the context for each segment and the worker yielded the caption for each segment under the guidance of the context. Xie \emph{et al.} \cite{xie2021hierarchical} proposed integrated recommendation on pictures and videos with the high-level agent as an item recommender and the low-level agent as a channel selector. For robotic manipulation, Yang \emph{et al.} \cite{yang2021hierarchical} trained the symbolic planing and kinematic control policies with hierarchical agents to enable the model to learn varied outcomes of the multi-step tasks. Gieselmann \emph{et al.} \cite{gieselmann2021planning} designed a planning algorithm for robots by dividing the original  Markov Decision Process (MDP) into a hierarchy of shorter horizon MDPs, and solved them with sub-goals at the bottom level of the hierarchy. In this paper, we generalize the hierarchical reinforcement learning framework for irregular object packing in order to effectively search the optimal sequence and placement in the large space.

\section{Approach}
\label{sec:method}
In this section, we first briefly introduce the problem of packing planning for irregular objects and the overall pipeline. Then we present the detail of the hierarchical reinforcement learning framework, where the high-level manager predicts the packing sequence and the low-level worker yields the placement including location and orientation for the selected objects. Finally, we formulate the reward function to train the agent in hierarchical reinforcement learning.

\subsection{Problem Statement and Overall Pipeline}
    The objective of packing planning for irregular objects is to predict the packing sequence and placement for each object according to the object shape, which provides the optimal object arrangement for robot manipulator to accomplish automatic packing. The main challenges for the problem are two-fold. First, high-resolution visual clues of irregular objects make significant contribution to generate optimal packing plans, and the visual perception module is required to represent the fine-grained visual information for arrangement prediction. Second, the packing sequence and placement are tightly coherent, and efficient search algorithms are necessary for simultaneously searching the optimal solution in the large composed space. To address these, we extract the informative visual feature maps from heightmaps of objects and boxes in different views, and search the sequence and placement for packing with the hierarchical reinforcement learning framework.
    
    Fig. \ref{fig:pipeline} demonstrates the overall pipeline of our framework. The {six principal view heightmaps} for each object from {3D models} and the packing box in the top-down view are leveraged to represent the visual information. The manager network selects the next object for packing according to the heightmaps of unpacked objects and the box heightmap. The heightmap of the selected object and the box heightmap are leveraged to yield the optimal position and orientation for object placement. The object selection and placement generation are iteratively implemented until all objects have been packed into the box or no space for unpacked items remains.

\subsection{Sequence Planning}
    The sequence planning module predicts the next object to pack according to the information of the packed objects in the box and all unpacked objects, since the arrangement of packed objects have significant influence on the choice of subsequent objects.
    
    \textbf{State:} The state space represents all unpacked objects and the current object arrangement in the box. The agent requires sufficient information from multiple viewpoints to demonstrate the visual clues of unpacked objects. We employ {six principal view heightmaps} of unpacked objects to illustrate the shape information for the manager agent. All {six principal view heightmaps} for unpacked objects are taken in the pose where objects are stable in the plane, since the {six principal view} orientations of stable poses exist in the object placement with higher possibility. Zero-planes of the heightmap in each viewpoint is the opposite plane of the object bounding box. Object arrangement in the packing box is depicted by the top-view heightmap.
    
    \textbf{Action: }The manager agent predicts the optimal object to be packed in the next placement based on the {six principal view heightmaps} of all unpacked objects and box heightmap. By concatenating the {six principal view heighmaps} of different unpacked objects with the box heightmap respectively, we acquire the high-resolution feature maps for placement of various unpacked objects via convolutional neural networks inspired by \cite{yang2018sgm}. The feature maps of all objects are concatenated to predict the score of being selected for placement, where the feature maps of packed objects are set to all-zero matrix to keep the dimension consistency for convolutional neural networks in score prediction. The unpacked object with the highest score is chosen to be packed into the box. After the placement is completed by the worker agent, the heightmap of the packing box is updated and the feature map for score prediction of the selected object is assigned with all-zero matrix. When the number of unpacked objects is extremely large, the manager agent select the object for packing from $K$ unpacked instances with the top-K bounding box volume for efficient search, since packing larger objects with higher priority results in better space utilization and placement stability.

\subsection{Placement Planning}
    The placement planning module predicts the placement position and orientation for the object selected by the sequence planning module. For each chosen object to be packed, the agent takes the action to place the instance into the box at predicted position $(X, Y)$ and orientation $(\phi, \theta, \psi)$, while $(X, Y)$ represent the horizontal position in $x$ and $y$ axis and $(\phi, \theta, \psi)$ are the Euler angles indicating roll, pitch and yaw. The vertical position $Z$ is determined by the lowest position of object placement due to the gravity, where the selected instance does not collide with the packed objects in the box.

    \textbf{State: } The state space represents the selected object to pack and the current object arrangement in the box. Similar to the sequence planning, the object arrangement in the packing box is demonstrated by the top-view heightmap. Since only top-down packing trajectories are allowed in the packing box for safety and stability, the selected object is prohibited to shift horizontally inside the box. Therefore, the top-view and height-view heightmaps of selected objects provide sufficient visual information for the chosen instance.

    \textbf{Action: }Inspired by \cite{zeng2018learning}, we first scan the selected object to acquire the top-view and bottom-view heightmaps in different roll and pitch, where the roll and pitch are discretized as $\mathcal{O}_{rp}=\{(\phi_1, \theta_1)\dots(\phi_n, \theta_n)\}$ in grids. We then rotate the above heightmaps to obtain the heightmaps in different yaw, where the yaw is quantized into $\mathcal{O}_y=\{0, \psi_1, \dots,\psi_m\}$ with the equal interval. The roll, pitch and yaw all range from $0$ to $2\pi$. Since the complexity of object scanning for heightmap acquisition is significantly reduced by heightmap rotation, the efficiency of placement planning is significantly enhanced during inference. We denote the top-view and bottom-view heightmaps of the selected object in the Euler angle $(\phi_i, \theta_i, \psi_j)$ as $H_t^{ij}$ and $H_b^{ij}$ respectively, which are concatenated to the box heightmap with crop and alignment to generate the optimal horizontal position for placement given the orientation. The placement score matrix $W^{ij}$ for the selected object in the Euler angle $(\phi_i, \theta_i, \psi_j)$ is yielded via convolutional neural networks, and the size of $W^{ij}$ keeps the same as the box heightmap. The element in the $x_{th}$ row and $y_{th}$ column of $W^{ij}$ is represented by $W^{ij}[x,y]$, which demonstrates the score  that the horizontal object center should be placed in the position $(x,y)$ of the box with the orientation $(\phi_i, \theta_i, \psi_j)$. The elements of the score matrix are set to zero if the corresponding placement is illegal, which include the marginal area with collision to the box and the position that the object exceeds box height after placement.
    The agent selects the orientation and horizontal position candidates with the highest score for placement, and calculate the vertical position $z$ for placement based on bottom-view heightmaps of the selected object and box heightmap \cite{Wang2019}:
    \begin{equation}
        z=\max_{s=-\lfloor w/2\rfloor}^{\lceil w/2\rceil-1}\max_{t=-\lfloor h/2\rfloor}^{\lceil h/2\rceil-1}(H_c[x+s, y+t]- H_b^{ij}[s, t])
    \end{equation}
    where $H_c[x,y]$ and $H_b^{ij}[x,y]$ mean the element in the $i_{th}$ row and $j_{th}$ column of $H_c$ and $H_b^{ij}$ respectively, and $w$ and $l$ are the width and length of the heightmaps. By placing the selected object with the position and orientation predicted by the placement planning module, the box space is effectively utilized for irregular object packing,

\subsection{Reward Function and Network Training}
	
    \begin{figure}
        \centering
        \includegraphics[width=0.95\linewidth]{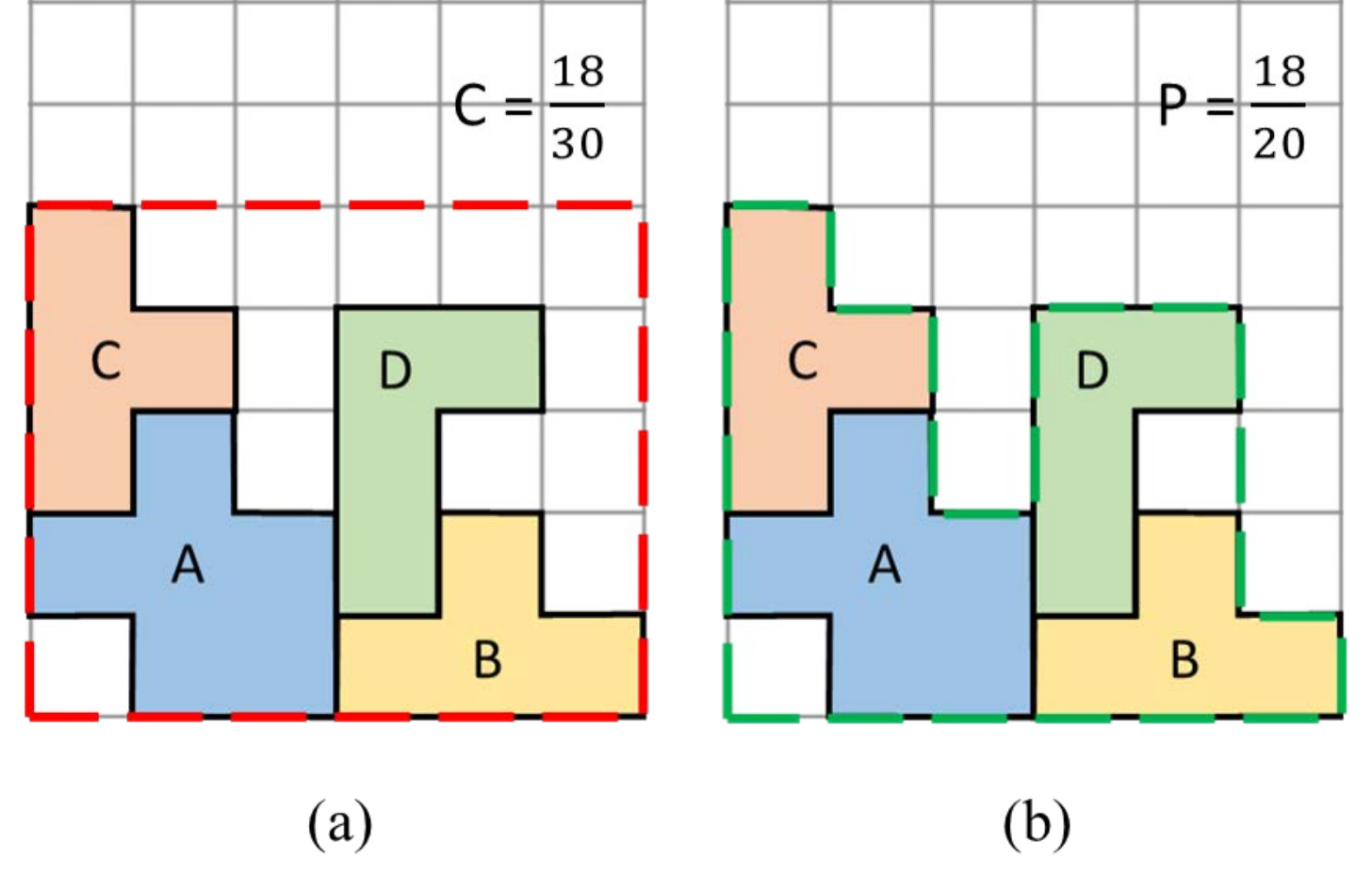}
        \caption{The objective function of compactness and pyramidality. (a) Compactness is the ratio between the total volume of packed objects and the minimum box size needed (red dashed line). (b) Pyramidality is the ratio between the total volume of packed objects and the region acquired by projecting all objects to the bottom of the box (green dashed line).}
        \vspace{-0.1cm}
        \label{fig:C_P}
     \end{figure}
     
    \begin{figure}[t!]
        \centering
        \subfloat[\label{fig:all}]
            {\includegraphics[width=0.167\textwidth]{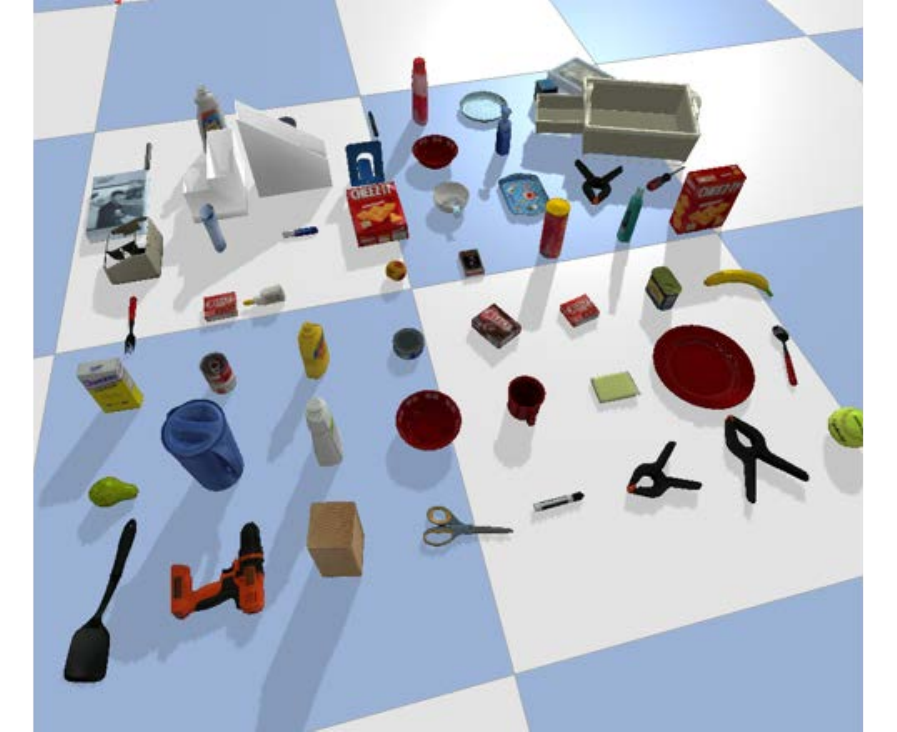}}
        \subfloat[\label{fig:easy}]
            {\includegraphics[width=0.155\textwidth]{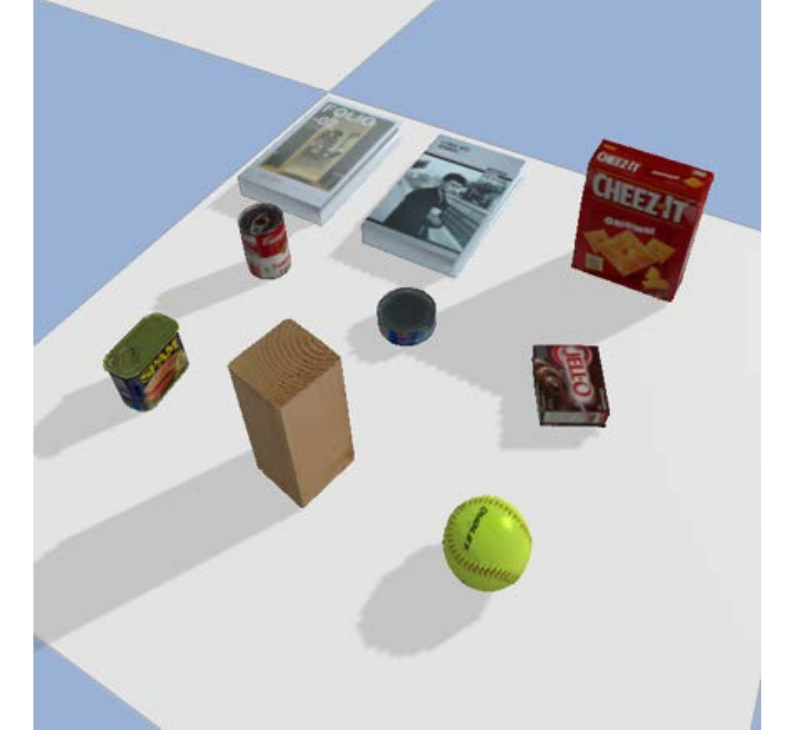}}
        \subfloat[\label{fig:hard}]
            {\includegraphics[width=0.16\textwidth]{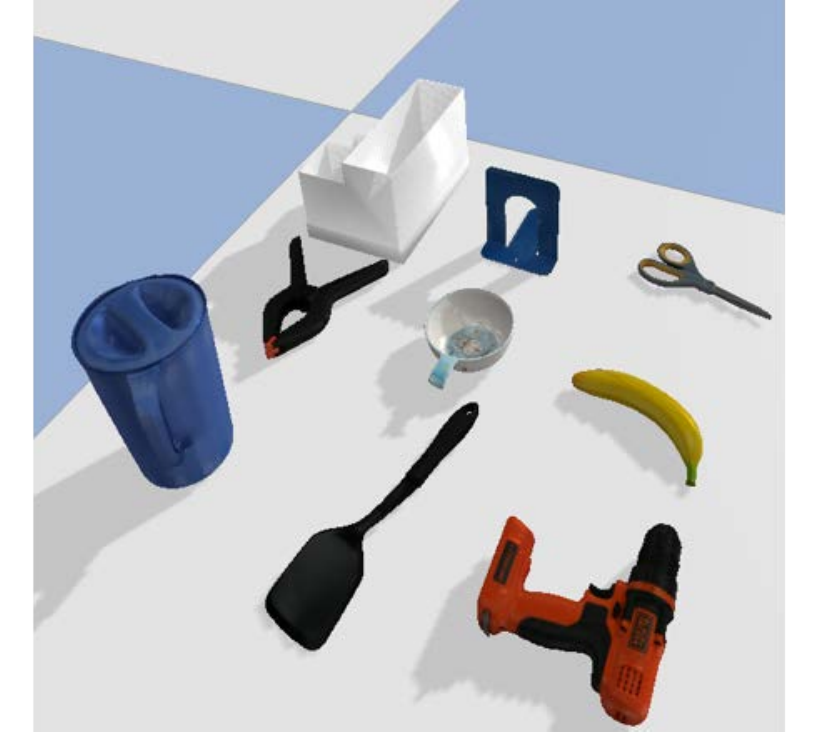}}
        \caption{(a) Part of select objects model fom YCB and OCRTOC datasets in our expriments. (b) and (c) show examples of object combinations in easy and hard experimental settings.}
        \label{fig:dataset}
        \vspace{-0.3cm}
    \end{figure}
	In order to effectively explore the search space, we utilize Q-Learning for the hierarchical reinforcement learning framework. Given state $s_t$ at the $t_{th}$ step for both the sequence and placement planning modules, the agents aim to take optimal action $a_t$ to iteratively maximize the expected reward function over the whole process. In robotic packing planning, our goal is to pack as many objects as possible into a packing box with fixed size. To achieve this goal, {we calculate the compactness $C$, the pyramidality $P$ \cite{hu2020tap} and the stability $S$ for the objective function}, which is formulated in the following for the state $s_t$:
    \begin{align}
    \max J(s_t)  & = \alpha C + \beta P + \gamma S\\& =\alpha \frac{\sum_{i=1}^tV_i}{LWh_t}+\beta \frac{\sum_{i=1}^tV_i}{V_p^t}+\gamma S
    \end{align}
    where $V_i$ means the volume of the $i_{th}$ packed objects in the box, and $L$ and $W$ are the length and width of the box respectively. $h_t$ represents the maximum height of packed objects in the box after putting the $t_{th}$ object into the container, which can be depicted by the largest value in the box heightmap. The volume for the projection of the $t$ packed objects to the bottom is represented by $V_p^t$, and can be obtained via the summation over the box heightmap. {$\alpha$, $\beta$ and $\gamma$ are hyperparameters to balance the importance of compactness, pyramidality and stability in the objective function. We detail the physical meanings of the compactness, pyramidality and stability in the objective function as follows}:

    \begin{enumerate}[(1)]
   \item The compactness $C$ is the ratio of occupied volume to minimum box volume that is capable of containing all the objects, as shown in Fig. \ref{fig:C_P}(a). To pack more instances given the height constraint, the cuboid space with the height of packed objects and with the base area of the box should be minimized to fully leverage the space in the box. Therefore, we encourage the agent to maximize the compactness to improve space utilization.
    \item The pyramidality $P$ is the ratio of occupied volume to the projected volume of packed objects to the bottom, as shown in Fig. \ref{fig:C_P}(b). Object arrangement in the box should leave more continuous space to be available for subsequent packing in order to enhance the space utilization. We encourage the agent to maximize the pyramidality to avoid continuous space being blocked by objects.
    \item {The stability factor $S$ is decided by comparing the planned placement with actual placement results in the simulated environment. We assign $S = 1$ for stable placements and $S = 0$ for unstable placements. The stable placement means the scenario that the position and orientation difference between the planned and the actual one in the simulated environment is smaller than a threshold.}
	\end{enumerate}

    We define the reward function for round $t$ based on the objective function as following:
    \begin{equation}
        r(s_t, a_t)=J(s_{t+1})-J(s_{t})
    \end{equation}
    where $r(s_t, a_t)$ represents the reward for the action $a_t$ in the state $s_t$, and the agent is encouraged to increase the objective function with the selected actions. To train the hierarchical framework efficiently, we employ a two-stage training method for faster convergence. First, we only train the placement planning module with heuristic packing sequence sorted by the bounding box volume. After obtaining a pre-trained agent for placement generation, we replace the heuristic method of sequence planing by the agent to jointly optimize the manager and the worker neural networks. During the parameter optimization, we update the policy at different time-scales for better cooperation between the high-level and low-level agents.

\section{Experiments}
\label{sec:experiments}
    In this section, we conduct extensive experiments in physical simulated environment (PyBullet). We first present the implementation details, including the workspace settings in the simulator, the model configuration and the dataset collection, and then introduce the evaluation metrics for robotic packing planning.  After that, we make a discussion for important hyperparameters in our methods, and compare our approach with state-of-the-art packing planning techniques to show the superiority of our framework brought by the sequence and placement planning module. {Finally, we deploy our packing planning method on a robot manipulator to verify the generalization ability in the real world.}

\begin{figure}[t!]
    \centering
    \subfloat[\label{fig:HM}]
        {\begin{minipage}[b]{0.155\textwidth}
        \includegraphics[width=\linewidth]{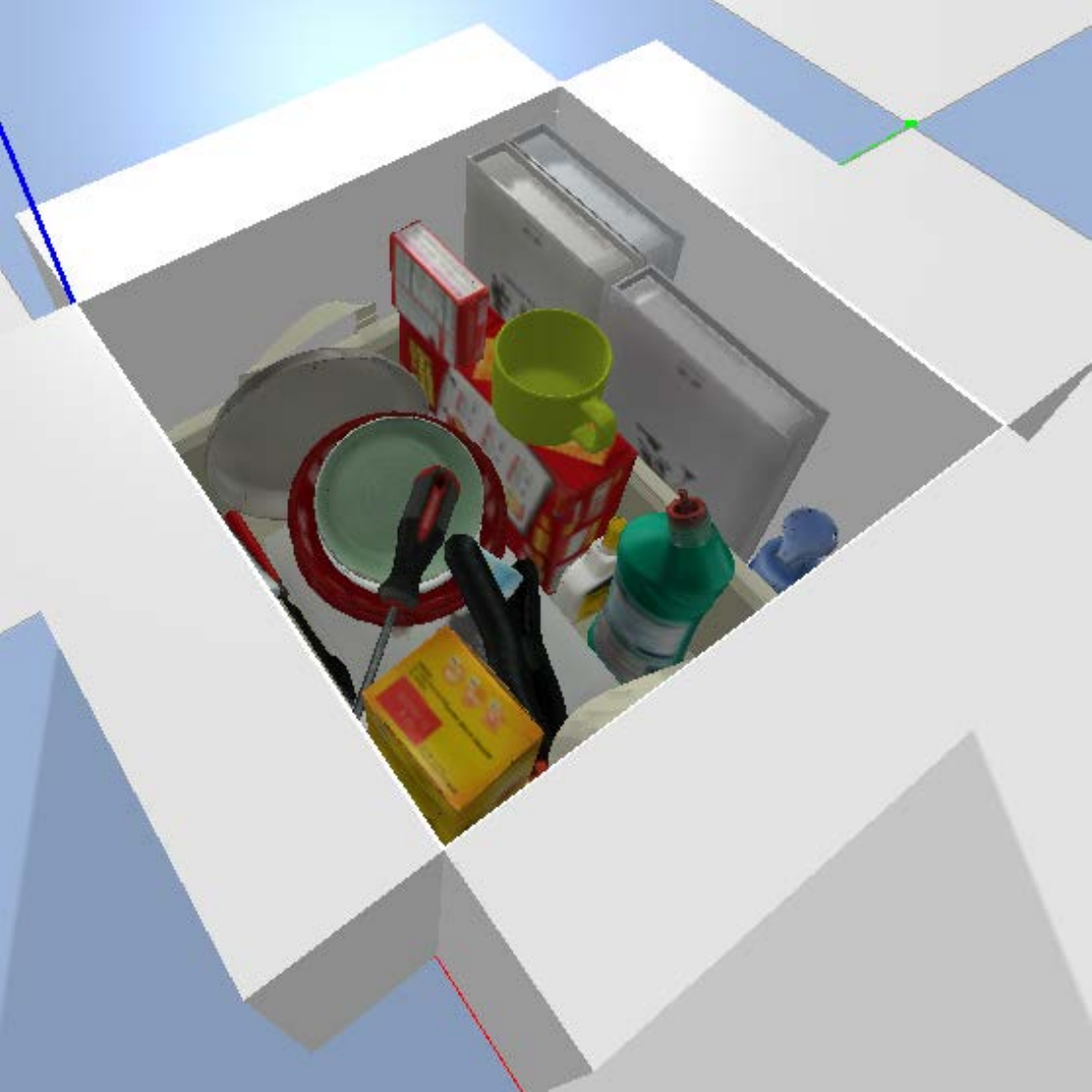}\vspace{5pt}
        \includegraphics[width=\linewidth]{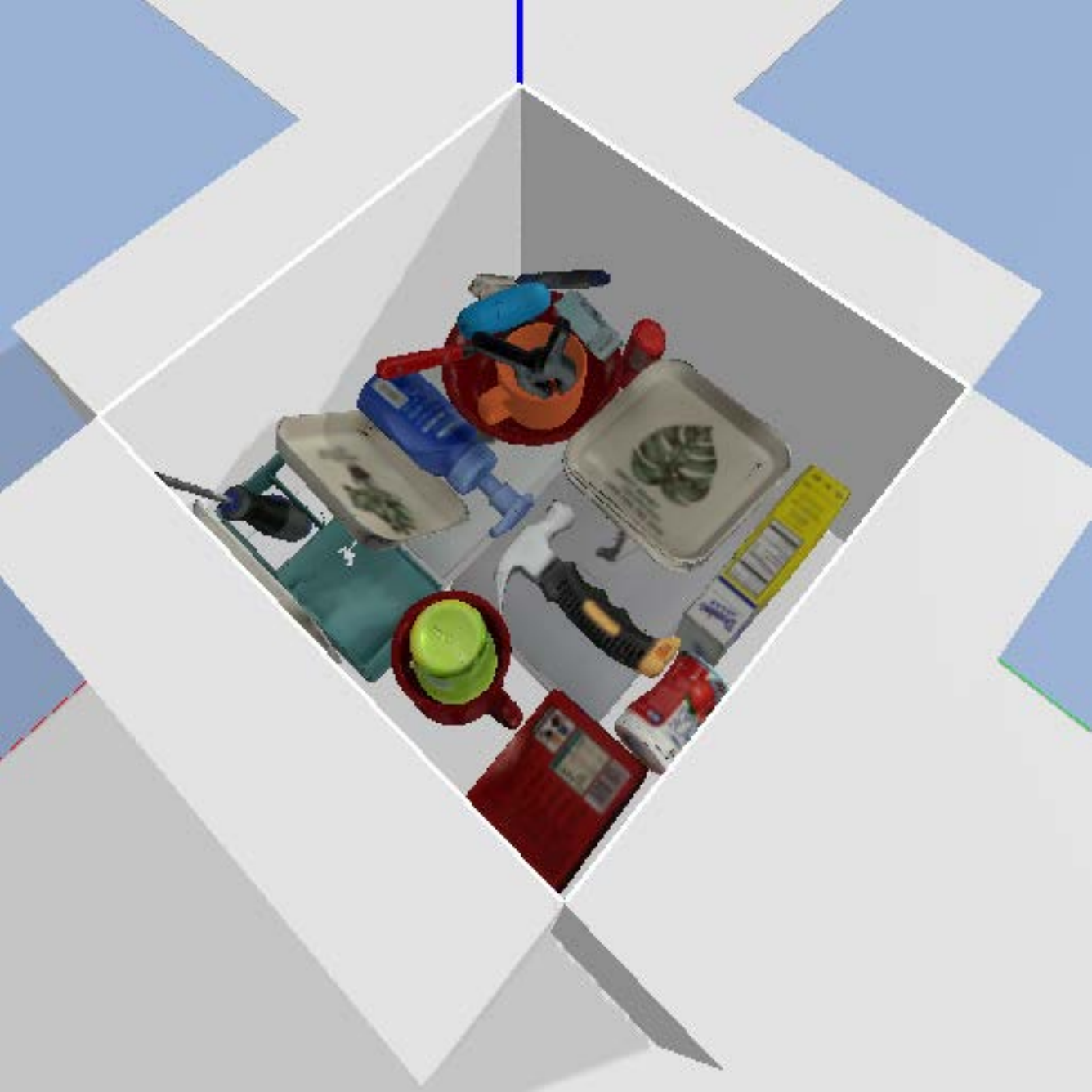}
        \end{minipage}
        }
    \subfloat[\label{fig:PP}]
        {\begin{minipage}[b]{0.155\textwidth}
        \includegraphics[width=\linewidth]{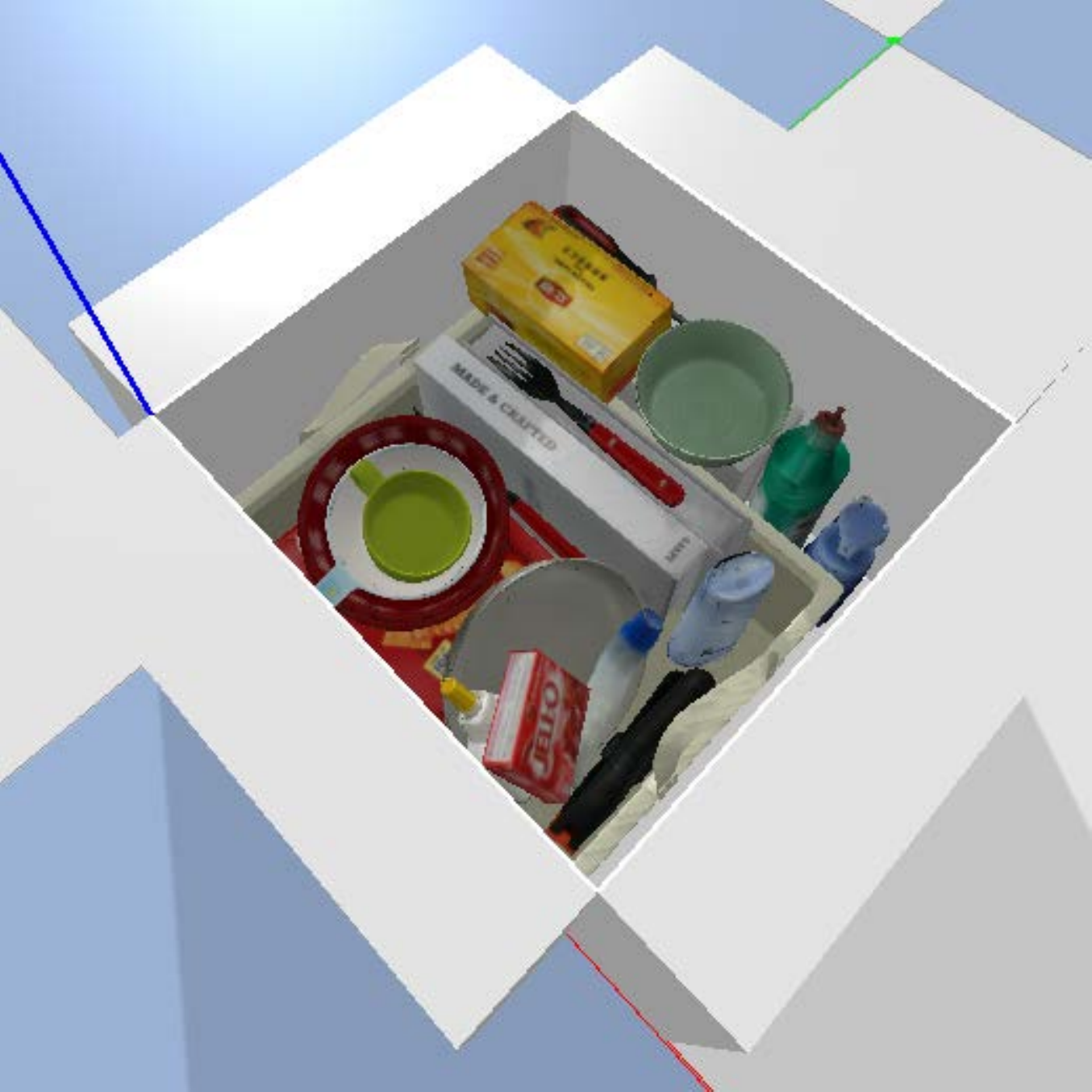}\vspace{5pt}
        \includegraphics[width=\linewidth]{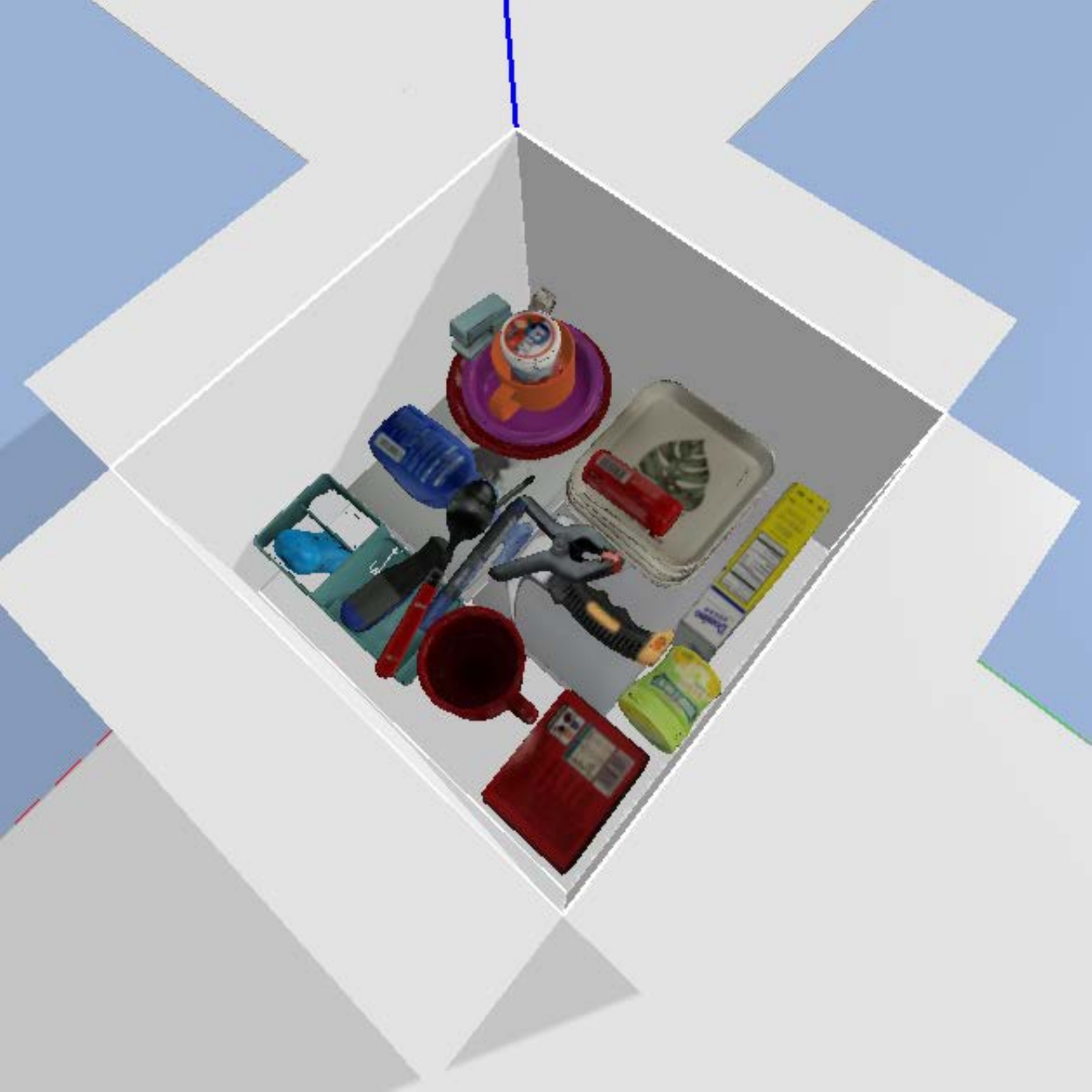}
        \end{minipage}
        }
    \subfloat[\label{fig:SP_PP}]
        {\begin{minipage}[b]{0.155\textwidth}
        \includegraphics[width=\linewidth]{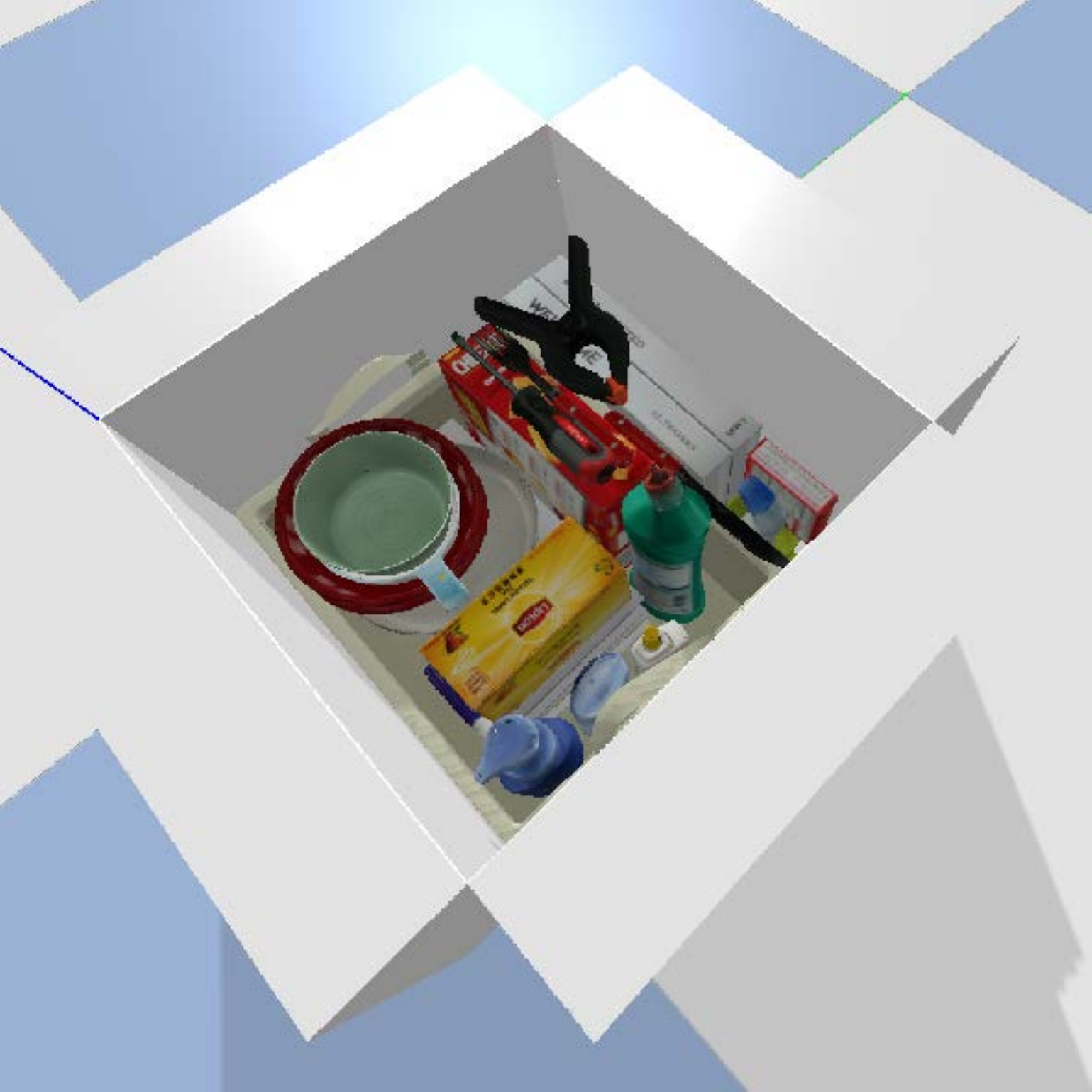}\vspace{5pt}
        \includegraphics[width=\linewidth]{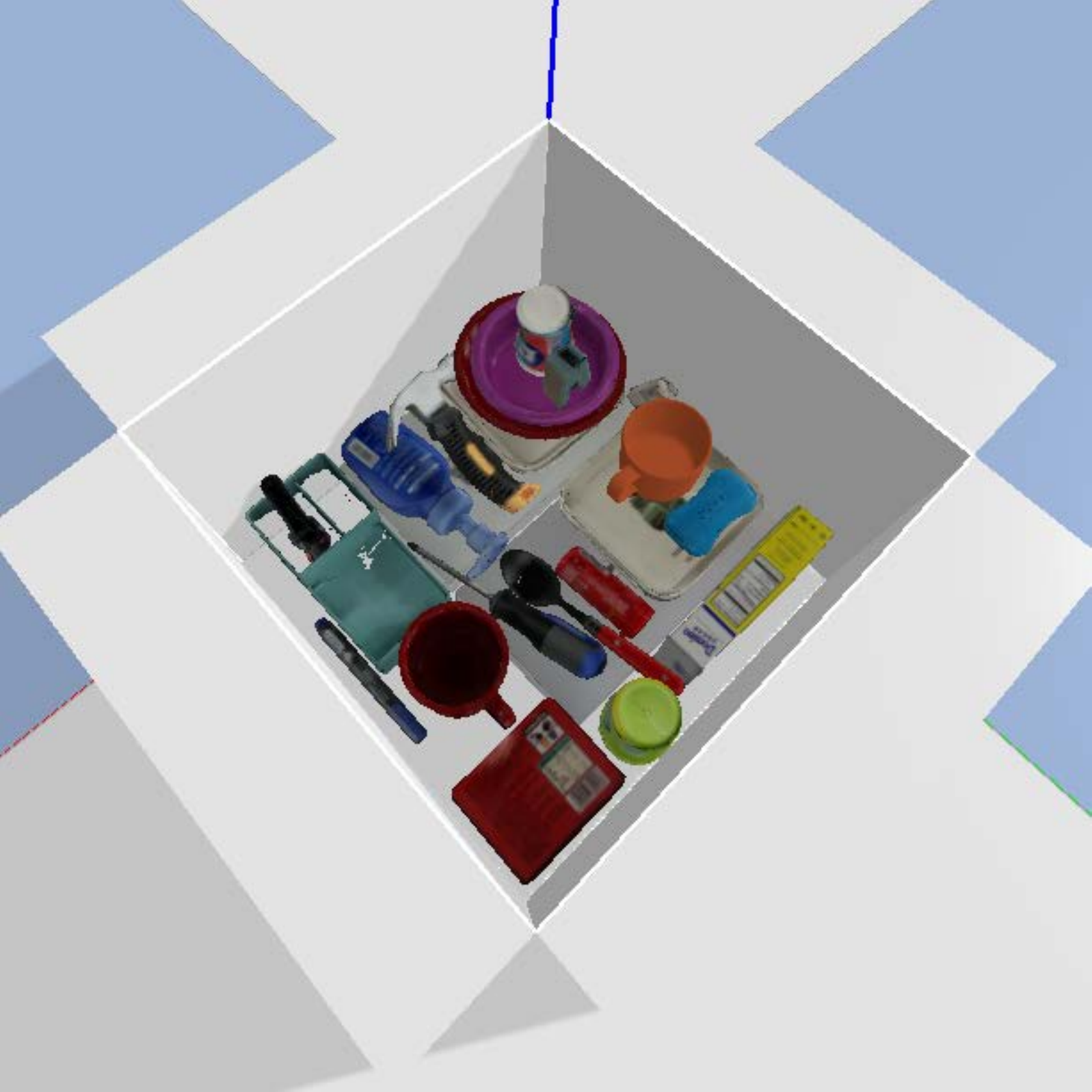}
        \end{minipage}
        }
    \caption{The visualization of packing plans obtained via (a) the B-Box Sequence + HM Heuristic, (b) the B-Box Sequence + Placement Planning and (c) the Sequence Planning + Placement Planning methods (ours), where our method learns to {stack bowls on plates to improve space utilization} and puts the screwdriver and the fork finally to improve stability.}
    \label{fig:visualization}
    \vspace{-0.5cm}
\end{figure}

\subsection{Implementation Details}

	In the simulated environment, we choose the packing box of size $40$cm$\times$40cm$\times$30cm as the constrained space for robotic packing. The spatial resolution of the RGB-D camera scanning the box heightmap is $200\times200$, which means each pixel in the heightmap represents a $2$mm$\times2$mm area in the base surface of the box. Heightmaps of objects are also scanned with the same resolution. For HM heuristic, we downsampled the box heightmap to $50\times50$.
	
	We leverage ResNet18\cite{he2016deep} as the backbone for visual feature extraction in the sequence planning module, which predicts the next object to pack via a three-layer fully-connected networks. In our experiments, we set $K=20$ as the maximum number of objects for policy generation of the sequence planning module. In the placement planning module, the search intervals for roll, pitch and yaw are all set as $\pi/2$ in default. We employ a U-Net \cite{ronneberger2015u} architecture network with 14 layers to generate the score matrix with the same size of box heightmap for optimal orientation and location selection of placement. {We utilize the Adam optimizer with the batchsize of 128. The learning rate is 1e-3 in the first stage and 1e-4 in the second stage. For hierarchical reinforcement learning, we update the placement planning agent for each epoch and the sequence planning agent per 4 epochs during the joint training stage. The hyperparameter $\alpha$, $\beta$ and $\gamma$ in the objective function are assigned with $0.75$, $0.25$, and $0.25$ respectively. The stability measurement thresholds for difference of positions and orientations are set to $2$cm and $\pi/6$.} 

	All object for packing in our experiments come from the YCB dataset \cite{calli2015benchmarking} and OCRTOC dataset \cite{liu2021ocrtoc}. We select 121 types of objects to construct our training and test dataset, as representative objects are shown in Fig. \ref{fig:all}. To evaluate the performance of our approach and the baseline methods in packing objects with different shapes, we discuss the performance variation in easy and hard experimental settings based on the shape regularity as visualized in Fig. \ref{fig:easy} and \ref{fig:hard}. Each case for robotic packing consists of 50 randomly selected instances, which are randomly scaled by 0.8$\times$ to 1.2$\times$ to improve the diversity. We collected 5,000 and 2,000 objects combinations as the training and test sets. All experiments are accelerated by an NVIDIA GeForce RTX 3090 GPU. We conducted agents training and simulation in parallel to avoid additional time cost during simulation.

\subsection{Evaluation Metrics}
	To evaluate the packing effectiveness, we utilize three metrics including the compactness, pyramidality, {stability} and average number of packed objects. Compactness evaluates space utilization of object arrangement, while pyramidality measures available space for subsequent object packing. {Stability demonstrates the difference between the planned object positions and orientation with the actual ones after placement. We also report the average time cost per object for efficiency evaluation.}

\subsection{Hyperparameter Discussions}
	In this section, we discuss the influence of important hyperparameters on our hierarchical reinforcement framework including the maximum number of objects $K$ for policy generation of the sequence planning module, the search intervals for placement orientation selection {and the resolution of heightmaps}.  According to the ablation study on easy cases, users can select the optimal hyperparameter settings based on the requirements including effectiveness and efficiency of packing.

\begin{table}[t]
    \centering
    \caption{{The performances of our method with different maximum number of objects $K$ for policy generation of sequence planning module in easy settings. \#Obj. means the number of packed objects and Lat. represents the latency for each object.}}
    \renewcommand\arraystretch{1.1}
    \begin{tabular}{p{2.1cm}<{\centering}|p{0.8cm}<{\centering}p{0.8cm}<{\centering}p{0.8cm}<{\centering}p{0.8cm}<{\centering}p{0.8cm}<{\centering}}
    \hline
    K & $C$ & $P$ & $S$ & \# Obj.& Lat.(s) \\ \hline
    0 & 0.451 & 0.758 & 0.827 & 40.07 & 0.62 \\ 
    10 & 0.454 & 0.763 & 0.834 & 40.25 & 0.80 \\
    20 & 0.456 & 0.766 & 0.836 & 40.38 & 0.97 \\
    30 & 0.457 & 0.766 & 0.837 & 40.41 & 1.16 \\\hline
    \end{tabular}
    \label{table:K}
\end{table}

\textbf{Discussion on maximum number $K$ in the sequence planning module:} The maximum number of objects $K$ for policy generation of the sequence planning module determines the size of search space during the sequence planning, because the number of unpacked objects is usually extremely large. To investigate the impact on packing effectiveness and efficiency, we set $K$ as 0 (same as sorted by bounding box size), 10, 20 and 30 respectively in our experiments with 50 objects for packing. Table \ref{table:K} shows the results, where we observe significant performance improvement with $K$ from 0 to 20 and slight enhancement with $K$ larger than 20. Therefore, We set $K$ to 20 in other experiments to achieve the satisfying trade-off between the effectiveness and efficiency.

\begin{table}[t!]
\centering
\caption{{The performances with different search intervals (SI) of orientations in easy settings.}}
    \renewcommand\arraystretch{1.1}
    \begin{tabular}{p{0.9cm}<{\centering}|p{1.1cm}<{\centering}|p{0.7cm}<{\centering}p{0.7cm}<{\centering}p{0.7cm}<{\centering}p{0.75cm}<{\centering}p{0.75cm}<{\centering}}
    \hline
    SI of $\psi$ & SI of $\phi$,$\theta$ &$C$ & $P$ & $S$ & \# Obj.& Lat.(s) \\ \hline
    \multirow{2}{*}{$\pi/2$} &$\pi/2$& 0.456 & 0.766 & 0.836 & 40.38 & 0.97 \\ 
    &$\pi/4$& 0.457 & 0.768 & 0.838 & 40.42 & 2.67 \\ \hline
    \multirow{2}{*}{$\pi/4$} &$\pi/2$& 0.457 & 0.767 & 0.836 & 40.41 & 1.51 \\ 
    &$\pi/4$& 0.458 & 0.769 & 0.838 & 40.46 & 4.49 \\
    \hline
    \end{tabular}
\label{table:SI}
\end{table}

\begin{table}[t]
    \centering
    \caption{{The performance in easy settings} for different input resolutions.}
    \renewcommand\arraystretch{1.1}
    \begin{tabular}{p{2.1cm}<{\centering}|p{0.8cm}<{\centering}p{0.8cm}<{\centering}p{0.8cm}<{\centering}p{0.8cm}<{\centering}p{0.8cm}<{\centering}}
    \hline
    Resolution & $C$ & $P$ & $S$ & \# Obj.& Lat.(s) \\ \hline
    100$\times$100 & 0.442 & 0.750 & 0.819 & 39.45 & 0.64 \\ 
    200$\times$200 & 0.456 & 0.766 & 0.836 & 40.38 & 0.97 \\
    400$\times$400 & 0.460 & 0.769 & 0.840 & 40.63 & 2.53  \\ \hline
    \end{tabular}
    \label{table:resolution}
\end{table}

\textbf{Discussion on search intervals of placement orientations:} To investigate the influence on packing effectiveness and efficiency, we set search intervals for the Euler angles including roll, pitch and yaw as $\pi/2$ and $\pi/4$ respectively. Table \ref{table:SI} demonstrates the results, where enlarging the search space for the Euler angles of placement orientation only slightly enhance the performance while significantly increases the latency. {Because most objects are stable in flat surface for initial orientations and the minimal interval between two consecutive stable orientations is $\pi/2$ for roll and pitch, the similar performance is observed where the search interval of $\pi/4$ and $\pi/2$ are employed respectively. The initial orientation in yaw usually enforces objects to nearly fit the box wall, and rotating the object in yaw with the intervals of $\pi/2$ can keep the property with high space utilization ratio. Therefore, changing the search interval from $\pi/4$ to $\pi/2$ does not significantly affect the performance. Objects with centrosymmetry in the above axis also obtain equivalence of different rotational angles.} Since the search space is exponentially amplified by the search intervals, we set the search intervals for raw, pitch and yaw as $\pi/2$ in other experiments to achieve the acceptable accuracy-complexity trade-off.

{\textbf{Discussion on resolution of heightmaps:} The resolution of heightmaps determines the size of search space during position selection. Lower resolution can cause performance loss due to the shrunk search space, while higher resolution costs more time in inference. To investigate the influence of input resolutions, we conduct experiments on different input resolutions. Table III shows the results, where the input resolution of 200$\times$200 outperforms that of 100$\times$100 in space utilization ratio and only slightly underperforms that of 400$\times$400 with significantly reduced latency. Therefore, we select the 200$\times$200 resolution in most experiments for better effectiveness-efficiency trade-offs.}

\subsection{Comparison with Baseline Methods}

    \textbf{Results on easy cases:} Objects with regular shapes such as boxes and cans are error-tolerant in packing, and those with irregular shapes such as power drills and scissors require extremely precise arrangement to avoid collision during placement. Therefore, we evaluate the packing planning methods in easy and hard experimental settings as visualized in Fig. \ref{fig:easy} and \ref{fig:hard}. The baseline methods are constructed in the following: replacing the sequence planning module (SP) by sorting the volume of bounding box for all objects (B-Box Seq) and substituting the placement planning module (PP) by the HM heuristic \cite{Wang2019}. {The PackIt-Heuristic method\cite{goyal2020packit} utilized volume order to select objects, Bottom-Left-Back Fill for position and dimension length order for orientation. The stable HM\cite{Wang2019} deployed strict force balance constraints on the basis of HM heuristic.} Table \ref{table:normal} demonstrates compactness, pyramidality, {stability}, average number of packed objects and latency per object for different methods in easy experiments. Compared to the baseline with heuristic sequence and placement algorithms, employing our placement planning module packs {increases the number of packed objects by 0.88 (40.07 vs. 39.19)} with higher compactness, pyramidality, stability and half of the latency. Utilizing our sequence planning module to replace HM heuristics also improves performance, as the sequence planning module prefers to pack instances with irregular shape later and guarantee evenness at lower levels in the box. Fusing the sequence and placement planning modules, our method outperforms the baseline by packs {1.19 more objects (40.38 vs. 39.19) with 0.30 seconds less latency per object (0.97s vs. 1.27s).} {Since we employ heightmaps for object representation without RGB textures, our method can be generalized to unseen objects without significantly performance dropping.} Fig. \ref{fig:visualization} shows an example for packing 25 irregular objects with different packing methods, where our method achieves the highest space utilization ratio. The parameter number of our agent is 10.23MB, and the training cost of our method is 16 GPU hours in average. 
\begin{table}[t]
    \centering
    \caption{{The compactness, pyramidality, stability, average number of packed objects and latency per object for different methods in easy settings.}}
    \renewcommand\arraystretch{1.1}
    \begin{tabular}{p{2.1cm}<{\centering}|p{0.8cm}<{\centering}p{0.8cm}<{\centering}p{0.8cm}<{\centering}p{0.8cm}<{\centering}p{0.8cm}<{\centering}}
    \hline
    Methods & $C$ & $P$ & $S$ & \# Obj.& Lat.(s) \\ \hline
    Random &  0.253  & 0.419 &  0.138 & 21.84 & - \\ \hline
    B-Box Seq + HM & 0.438 & 0.746 & 0.716 & 39.19 & 1.27\\
    B-Box Seq + PP & 0.451 & 0.758 & 0.827 & 40.07  & 0.62\\
    SP + HM & 0.442 & 0.752 & 0.724 & 39.43 & 1.63\\ \hline
    PackIt-Heuristic & 0.417 & 0.712 & 0.663 & 35.28 & 0.17\\ 
    Stable HM\cite{Wang2019} & 0.443 & 0.753 & 0.882 & 39.49 &2.38 \\ 
    SP + PP (Ours) & 0.456 & 0.766 & 0.836 & 40.38 & 0.97 \\ \hline
    \end{tabular}
    \vspace{-0.1cm}
    \label{table:normal}
\end{table}
\begin{table}[t]
    \centering
    \caption{{The compactness, pyramidality, stability, average number of packed objects and latency per object for different methods in hard settings.}}
    \renewcommand\arraystretch{1.1}
    \begin{tabular}{p{2.1cm}<{\centering}|p{0.8cm}<{\centering}p{0.8cm}<{\centering}p{0.8cm}<{\centering}p{0.8cm}<{\centering}p{0.8cm}<{\centering}}
    \hline
    Methods & $C$ & $P$ & $S$ & \# Obj.& Lat.(s) \\ \hline
    Random &  0.216  & 0.382 & 0.104  & 19.33 & - \\ \hline
    B-Box Seq + HM & 0.385 & 0.673 & 0.678 & 34.82 & 1.59\\
    B-Box Seq + PP & 0.415 & 0.703 & 0.760 & 36.26 & 0.63\\
    SP + HM & 0.392 & 0.676 & 0.683  & 34.98 & 2.05\\ \hline
    PackIt-Heuristic & 0.367 & 0.643 & 0.649 & 32.48 & 0.17 \\ 
    Stable HM\cite{Wang2019} & 0.397 & 0.695 & 0.834 & 35.61 & 2.60 \\ 
    SP + PP (Ours) & 0.423 & 0.716 & 0.775 & 37.28 & 0.97\\ \hline
    \end{tabular}
    \vspace{-0.3cm}
    \label{table:hard}
\end{table}

    \textbf{Results on hard cases:} The heightmap resolution for heuristic packing methods is limited due to the high computational complexity, which performs poorly on the space utilization and collision prevention for irregular object packing. On the contrary, our method directly generate the optimal packing sequence and placement by one forward pass, and the fine-grained visual clues represented by high-resolution images can be considered efficiently. Table \ref{table:hard} shows the performance of different methods in hard experimental settings. Our framework outperforms more significantly for hard cases on effectiveness compared with the easy experimental settings, because the irregular object packing is sensitive to placement position and the space utilization ratio is affected by the arrangement more obviously. Moreover, HM usually degrades into grid search on higher space dimensions in irregular object packing, so that our method significantly decreases the average time cost.

\begin{figure}[t!]
    \centering
    \subfloat[\label{fig:Platform}]
        {\includegraphics[width=0.140\textwidth]{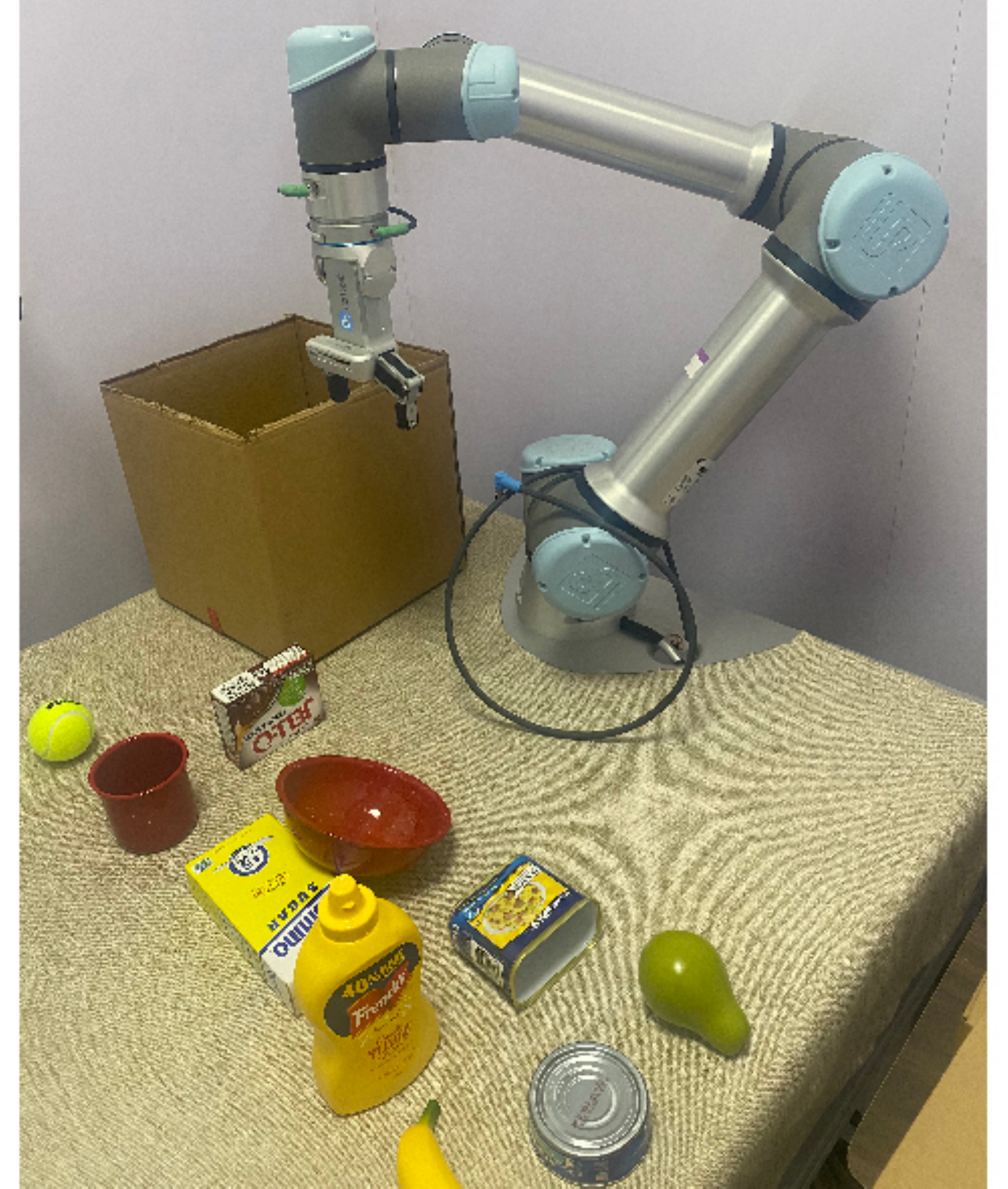}}
    \subfloat[\label{fig:Plan}]
        {\includegraphics[width=0.175\textwidth]{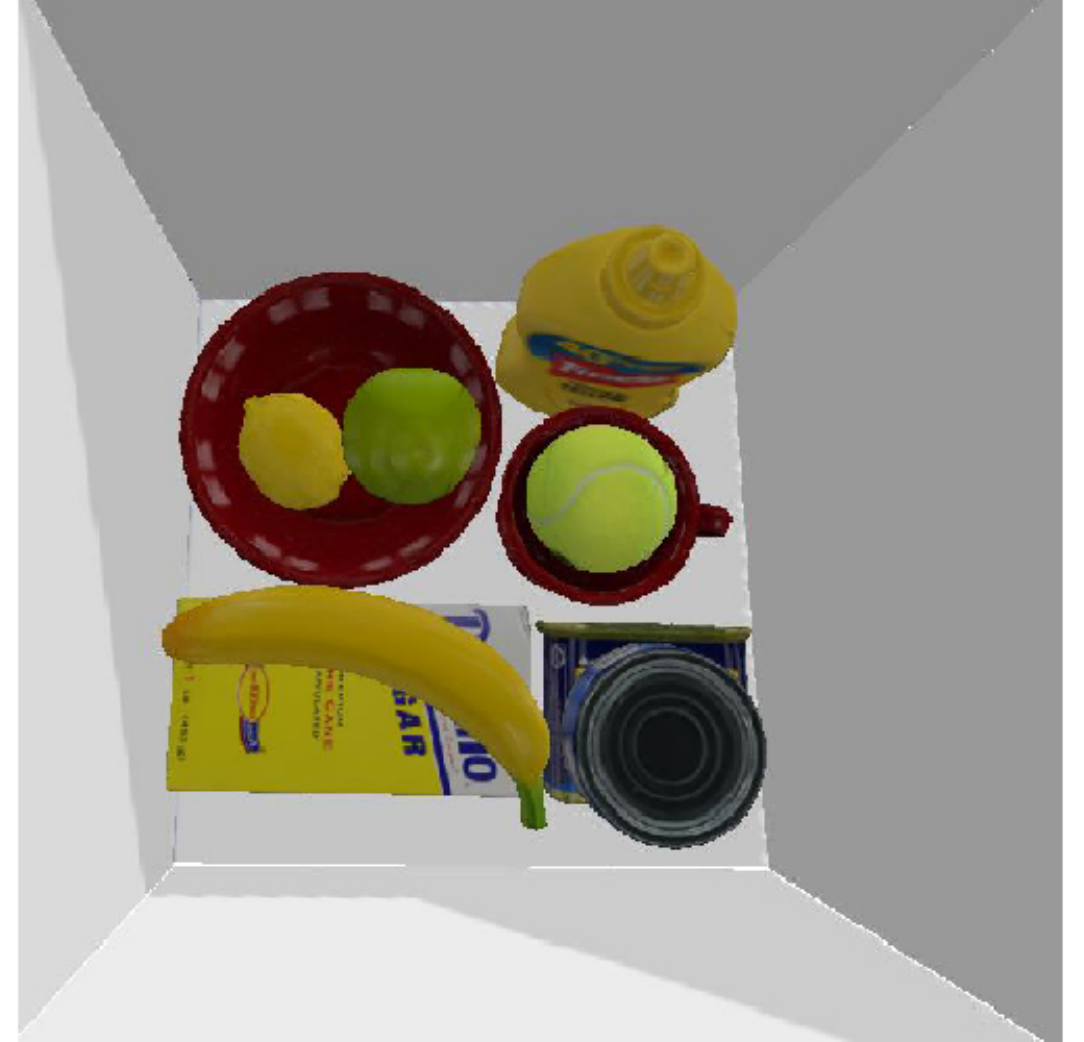}}
    \subfloat[\label{fig:Real}]
        {\includegraphics[width=0.175\textwidth]{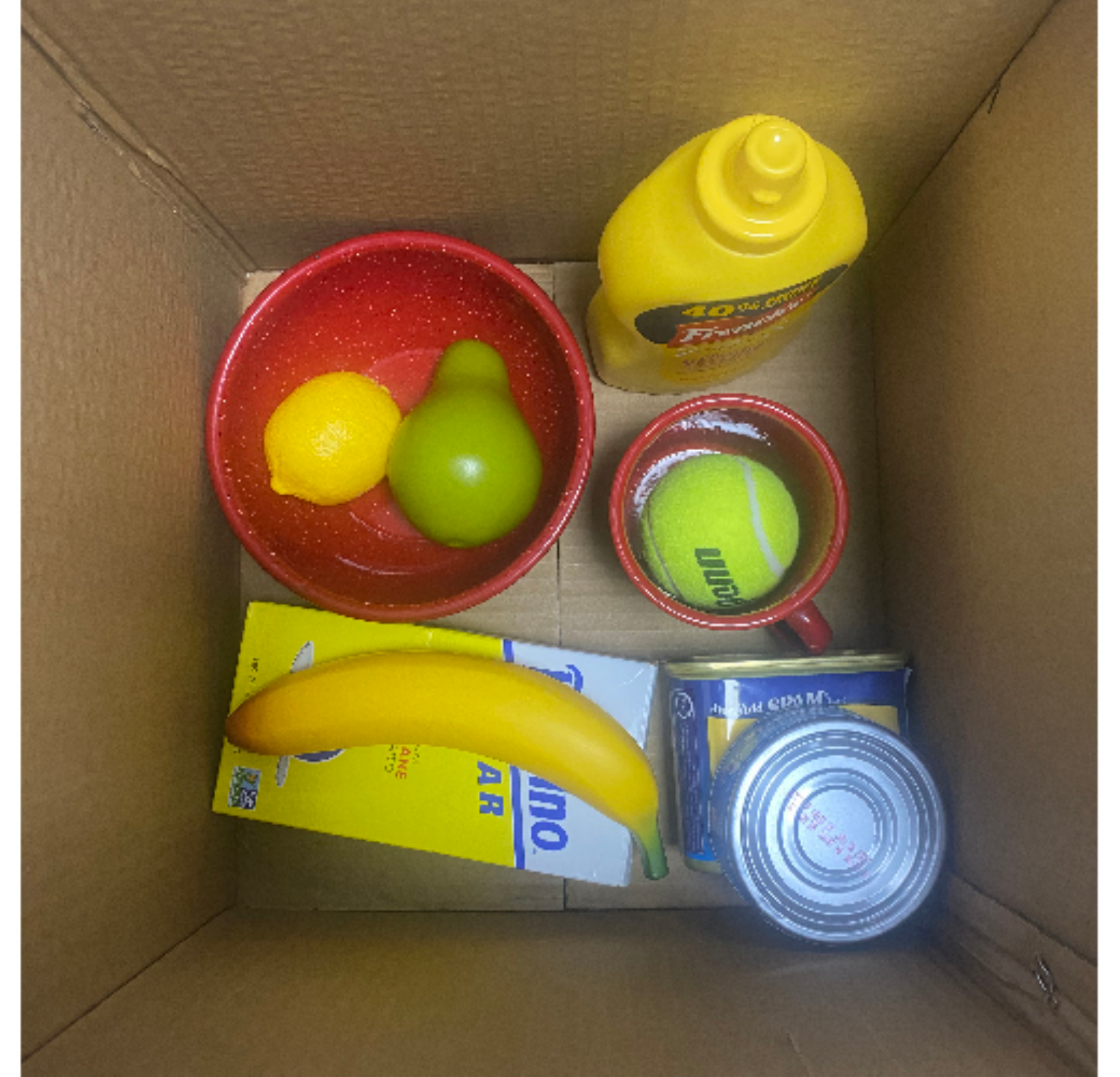}}
        \vspace{0.1cm}
    \caption{{The physical experiment of our method. (a) Our experimental packing setup. The grasp and placement are implemented by a UR5 robot equipped with a parallel-jaw gripper. (b) and (c) are examples of a 10-item placement in simulation and real world.}}
    \label{fig:Physical}
    \vspace{-0.3cm}
\end{figure}

\subsection{Experiments on Physical Platform}
{
    To verify the feasibility of our method on real robot manipulators, we conducted experiments for 10-item packing problem on known objects. As shown in Fig. \ref{fig:Physical}(a), we utilize a UR5 robot equipped with a parallel-jaw RG2 gripper for grasp and placement. We employ three RealSense D435 depth camera, including one on UR5 to obtain box heightmaps and two observing workspace for object perception. The packing box size is 30cm$\times$30cm$\times$30cm, and the items are from the YCB object dataset. Categories and geometry details of objects are captured by the two cameras observing the
    workspace with perception module based on \cite{wu2022smart}. We generated the six principal view heightmaps
    of objects in the simulator based on estimated categories and deformations, and obtained the heightmap of box through the camera on the UR5 arm. After sequence and placement planing, we leverage models in \cite{liu2022ge} to generate grasp and utilize the top-down path for placement. Experiments show that our method can be implemented by real robot manipulators and the objects can be packed as planned. The demonstration video is provided in the supplementary material.}


\section{Conclusion and Limitations}
\label{sec:conclusion}
In this paper, we have proposed a deep hierarchical reinforcement learning approach to simultaneously generate packing sequence and placement arrangement for irregular object packing planning. The sequence planning module extracts features from {six principal view heightmaps} of all instances to predict the next object for packing, and the placement planning module yields the optimal position and orientation without collision and space waste for the selected object. Extensive experiments demonstrates the effectiveness of our approach.
    
The limitations of our approach can be summarized as follows. Representing 3D shape with 2D heightmaps requires multiple viewpoints, and causes high computational cost and information loss during inference. {When deploying our method in real world, the perception errors of object shape and the placement infeasibility may decrease the space utilization ratio and even cause the package to fail. Therefore, enhancing the computational efficiency and the robustness in real-world deployment will be interesting future works.}


\bibliography{packing}

\begin{thebibliography}{10}
\providecommand{\url}[1]{#1}
\csname url@rmstyle\endcsname
\providecommand{\newblock}{\relax}
\providecommand{\bibinfo}[2]{#2}
\providecommand\BIBentrySTDinterwordspacing{\spaceskip=0pt\relax}
\providecommand\BIBentryALTinterwordstretchfactor{4}
\providecommand\BIBentryALTinterwordspacing{\spaceskip=\fontdimen2\font plus
\BIBentryALTinterwordstretchfactor\fontdimen3\font minus
  \fontdimen4\font\relax}
\providecommand\BIBforeignlanguage[2]{{%
\expandafter\ifx\csname l@#1\endcsname\relax
\typeout{** WARNING: IEEEtran.bst: No hyphenation pattern has been}%
\typeout{** loaded for the language `#1'. Using the pattern for}%
\typeout{** the default language instead.}%
\else
\language=\csname l@#1\endcsname
\fi
#2}}

\bibitem{Wang2019}
F.~Wang and K.~Hauser, ``Stable bin packing of non-convex 3d objects with a
  robot manipulator,'' in \emph{ICRA}, 2019, pp. 8698--8704.

\bibitem{Tanaka2020}
T.~Tanaka, T.~Kaneko, M.~Sekine, V.~Tangkaratt, and M.~Sugiyama, ``Simultaneous
  planning for item picking and placing by deep reinforcement learning,'' in
  \emph{IROS}, 2020.

\bibitem{Karabulut2004}
K.~Karabulut and M.~M. {\.I}nceo{\u{g}}lu, ``A hybrid genetic algorithm for
  packing in 3d with deepest bottom left with fill method,'' in
  \emph{International Conference on Advances in Information Systems}, 2004, pp.
  441--450.

\bibitem{Zhao2021Advanced}
Y.~Zhao, C.~Rausch, and C.~Haas, ``Optimizing 3d irregular object packing from
  3d scans using metaheuristics,'' \emph{Advanced Engineering Informatics},
  vol.~47, p. 101234, 2021.

\bibitem{Hu2017}
H.~Hu, X.~Zhang, X.~Yan, L.~Wang, and Y.~Xu, ``Solving a new 3d bin packing
  problem with deep reinforcement learning method,'' \emph{arXiv preprint
  arXiv:1708.05930}, 2017.

\bibitem{Zhao2021}
H.~Zhao, Q.~She, C.~Zhu, Y.~Yang, and K.~Xu, ``Online 3d bin packing with
  constrained deep reinforcement learning,'' in \emph{AAAI}, vol.~35, no.~1,
  2021, pp. 741--749.

\bibitem{Chazelle1989}
B.~Chazelle, H.~Edelsbrunner, and L.~J. Guibas, ``The complexity of cutting
  complexes,'' \emph{Discrete \& Computational Geometry}, vol.~4, no.~2, pp.
  139--181, 1989.

\bibitem{Iori2021}
M.~Iori, V.~L. de~Lima, S.~Martello, F.~K. Miyazawa, and M.~Monaci, ``Exact
  solution techniques for two-dimensional cutting and packing,'' \emph{European
  Journal of Operational Research}, vol. 289, no.~2, pp. 399--415, 2021.

\bibitem{che2021machine}
Y.~Che, K.~Hu, Z.~Zhang, and A.~Lim, ``Machine scheduling with orientation
  selection and two-dimensional packing for additive manufacturing,''
  \emph{Computers \& Operations Research}, vol. 130, p. 105245, 2021.

\bibitem{abdel2018improved}
M.~Abdel-Basset, G.~Manogaran, L.~Abdel-Fatah, and S.~Mirjalili, ``An improved
  nature inspired meta-heuristic algorithm for 1-d bin packing problems,''
  \emph{Personal and Ubiquitous Computing}, vol.~22, no.~5, pp. 1117--1132,
  2018.

\bibitem{Ali2022}
S.~Ali, A.~G. Ramos, M.~A. Carravilla, and J.~F. Oliveira, ``On-line
  three-dimensional packing problems: a review of off-line and on-line solution
  approaches,'' \emph{Computers \& Industrial Engineering}, pp. 108--122, 2022.

\bibitem{zhang2021attend2pack}
J.~Zhang, B.~Zi, and X.~Ge, ``Attend2pack: Bin packing through deep
  reinforcement learning with attention,'' \emph{arXiv preprint
  arXiv:2107.04333}, 2021.

\bibitem{mazyavkina2021reinforcement}
N.~Mazyavkina, S.~Sviridov, S.~Ivanov, and E.~Burnaev, ``Reinforcement learning
  for combinatorial optimization: A survey,'' \emph{Computers \& Operations
  Research}, vol. 134, p. 105400, 2021.

\bibitem{Jangiti2018}
S.~Jangiti and S.~S. VS, ``Scalable and direct vector bin-packing heuristic
  based on residual resource ratios for virtual machine placement in cloud data
  centers,'' \emph{Computers \& Electrical Engineering}, vol.~68, pp. 44--61,
  2018.

\bibitem{El2019}
W.~H. El-Ashmawi and D.~S. Abd~Elminaam, ``A modified squirrel search algorithm
  based on improved best fit heuristic and operator strategy for bin packing
  problem,'' \emph{Applied Soft Computing}, vol.~82, p. 105565, 2019.

\bibitem{Bortfeldt1998}
A.~Bortfeldt and H.~Gehring, ``Applying tabu search to container loading
  problems,'' in \emph{Operations Research Proceedings 1997}, 1998, pp.
  533--538.

\bibitem{Faroe2003}
O.~Faroe, D.~Pisinger, and M.~Zachariasen, ``Guided local search for the
  three-dimensional bin-packing problem,'' \emph{Informs journal on computing},
  vol.~15, no.~3, pp. 267--283, 2003.

\bibitem{Ramos2016}
A.~G. Ramos, J.~F. Oliveira, J.~F. Gon{\c{c}}alves, and M.~P. Lopes, ``A
  container loading algorithm with static mechanical equilibrium stability
  constraints,'' \emph{Transportation Research Part B: Methodological},
  vol.~91, pp. 565--581, 2016.

\bibitem{Zhao2019}
Y.~Zhao and C.~T. Haas, ``A 3d irregular packing algorithm using point cloud
  data,'' in \emph{Computing in Civil Engineering 2019: Data, Sensing, and
  Analytics}, 2019, pp. 201--208.

\bibitem{goyal2020packit}
A.~Goyal and J.~Deng, ``Packit: A virtual environment for geometric planning,''
  in \emph{ICML}, 2020, pp. 3700--3710.

\bibitem{kulkarni2016hierarchical}
T.~D. Kulkarni, K.~Narasimhan, A.~Saeedi, and J.~Tenenbaum, ``Hierarchical deep
  reinforcement learning: Integrating temporal abstraction and intrinsic
  motivation,'' in \emph{NeurIPS}, 2016, pp. 3675--3683.

\bibitem{vezhnevets2017feudal}
A.~S. Vezhnevets, S.~Osindero, T.~Schaul, N.~Heess, M.~Jaderberg, D.~Silver,
  and K.~Kavukcuoglu, ``Feudal networks for hierarchical reinforcement
  learning,'' in \emph{ICML}, 2017, pp. 3540--3549.

\bibitem{nachum2018data}
O.~Nachum, S.~S. Gu, H.~Lee, and S.~Levine, ``Data-efficient hierarchical
  reinforcement learning,'' in \emph{NeurIPS}, 2018, pp. 3303--3313.

\bibitem{sutton1998intra}
R.~S. Sutton, D.~Precup, and S.~P. Singh, ``Intra-option learning about
  temporally abstract actions,'' in \emph{ICML}, vol.~98, 1998, pp. 556--564.

\bibitem{florensa2017stochastic}
C.~Florensa, Y.~Duan, and P.~Abbeel, ``Stochastic neural networks for
  hierarchical reinforcement learning,'' \emph{arXiv preprint
  arXiv:1704.03012}, 2017.

\bibitem{eppe2022intelligent}
M.~Eppe, C.~Gumbsch, M.~Kerzel, P.~D. Nguyen, M.~V. Butz, and S.~Wermter,
  ``Intelligent problem-solving as integrated hierarchical reinforcement
  learning,'' \emph{Nature Machine Intelligence}, vol.~4, no.~1, pp. 11--20,
  2022.

\bibitem{wang2020learning}
Z.~Wang, J.~Lu, and J.~Zhou, ``Learning channel-wise interactions for binary
  convolutional neural networks,'' \emph{IEEE Transactions on Pattern Analysis
  and Machine Intelligence}, vol.~43, no.~10, pp. 3432--3445, 2020.

\bibitem{wang2018video}
X.~Wang, W.~Chen, J.~Wu, Y.-F. Wang, and W.~Yang~Wang, ``Video captioning via
  hierarchical reinforcement learning,'' in \emph{CVPR}, 2018, pp. 4213--4222.

\bibitem{zhao2017multiresolution}
D.~Zhao, Y.~Ma, Z.~Jiang, and Z.~Shi, ``Multiresolution airport detection via
  hierarchical reinforcement learning saliency model,'' \emph{IEEE Journal of
  Selected Topics in Applied Earth Observations and Remote Sensing}, vol.~10,
  no.~6, pp. 2855--2866, 2017.

\bibitem{kim2021landmark}
J.~Kim, Y.~Seo, and J.~Shin, ``Landmark-guided subgoal generation in
  hierarchical reinforcement learning,'' \emph{NeurIPS}, vol.~34, 2021.

\bibitem{yang2021hierarchical}
X.~Yang, Z.~Ji, J.~Wu, Y.-K. Lai, C.~Wei, G.~Liu, and R.~Setchi, ``Hierarchical
  reinforcement learning with universal policies for multistep robotic
  manipulation,'' \emph{IEEE Transactions on Neural Networks and Learning
  Systems}, 2021.

\bibitem{lampinen2021towards}
A.~Lampinen, S.~Chan, A.~Banino, and F.~Hill, ``Towards mental time travel: a
  hierarchical memory for reinforcement learning agents,'' \emph{NeurIPS},
  vol.~34, 2021.

\bibitem{wan2021reasoning}
G.~Wan, S.~Pan, C.~Gong, C.~Zhou, and G.~Haffari, ``Reasoning like human:
  Hierarchical reinforcement learning for knowledge graph reasoning,'' in
  \emph{IJCAI}, 2021, pp. 1926--1932.

\bibitem{xie2021hierarchical}
R.~Xie, S.~Zhang, R.~Wang, F.~Xia, and L.~Lin, ``Hierarchical reinforcement
  learning for integrated recommendation,'' in \emph{AAAI}, vol.~35, no.~5,
  2021, pp. 4521--4528.

\bibitem{gieselmann2021planning}
R.~Gieselmann and F.~T. Pokorny, ``Planning-augmented hierarchical
  reinforcement learning,'' \emph{RAL}, vol.~6, no.~3, pp. 5097--5104, 2021.

\bibitem{yang2018sgm}
P.~Yang, X.~Sun, W.~Li, S.~Ma, W.~Wu, and H.~Wang, ``Sgm: sequence generation
  model for multi-label classification,'' \emph{arXiv preprint
  arXiv:1806.04822}, 2018.

\bibitem{zeng2018learning}
A.~Zeng, S.~Song, S.~Welker, J.~Lee, A.~Rodriguez, and T.~Funkhouser,
  ``Learning synergies between pushing and grasping with self-supervised deep
  reinforcement learning,'' in \emph{IROS}, 2018, pp. 4238--4245.

\bibitem{hu2020tap}
R.~Hu, J.~Xu, B.~Chen, M.~Gong, H.~Zhang, and H.~Huang, ``Tap-net:
  transport-and-pack using reinforcement learning,'' \emph{ACM Transactions on
  Graphics (TOG)}, vol.~39, no.~6, pp. 1--15, 2020.

\bibitem{he2016deep}
K.~He, X.~Zhang, S.~Ren, and J.~Sun, ``Deep residual learning for image
  recognition,'' in \emph{CVPR}, 2016, pp. 770--778.

\bibitem{ronneberger2015u}
O.~Ronneberger, P.~Fischer, and T.~Brox, ``U-net: Convolutional networks for
  biomedical image segmentation,'' in \emph{International Conference on Medical
  image computing and computer-assisted intervention}, 2015, pp. 234--241.

\bibitem{calli2015benchmarking}
B.~Calli, A.~Walsman, A.~Singh, S.~Srinivasa, P.~Abbeel, and A.~M. Dollar,
  ``Benchmarking in manipulation research: The ycb object and model set and
  benchmarking protocols,'' \emph{arXiv preprint arXiv:1502.03143}, 2015.

\bibitem{liu2021ocrtoc}
Z.~Liu, W.~Liu, Y.~Qin, F.~Xiang, M.~Gou, S.~Xin, M.~A. Roa, B.~Calli, H.~Su,
  Y.~Sun, \emph{et~al.}, ``Ocrtoc: A cloud-based competition and benchmark for
  robotic grasping and manipulation,'' \emph{IEEE Robotics and Automation
  Letters}, vol.~7, no.~1, pp. 486--493, 2021.

\bibitem{wu2022smart}
Z.~Wu, Z.~Wang, Z.~Wei, Y.~Wei, and H.~Yan, ``Smart explorer: Recognizing
  objects in dense clutter via interactive exploration,'' \emph{arXiv preprint
  arXiv:2208.03496}, 2022.

\bibitem{liu2022ge}
Z.~Liu, Z.~Wang, S.~Huang, J.~Zhou, and J.~Lu, ``Ge-grasp: Efficient
  target-oriented grasping in dense clutter,'' \emph{arXiv preprint
  arXiv:2207.11941}, 2022.

\end{thebibliography}

\end{document}